\documentclass{article}

\usepackage{PRIMEarxiv}

\usepackage[utf8]{inputenc} 
\usepackage[T1]{fontenc}    
\usepackage{hyperref}       
\usepackage{url}            
\usepackage{booktabs}    
\usepackage{tablefootnote}

\usepackage{amsfonts}       
\usepackage{nicefrac}       
\usepackage{comment}
\usepackage{microtype}      
\usepackage{lipsum}
\usepackage{xcolor}
\usepackage{algpseudocode}
\usepackage[ruled,vlined]{algorithm2e}
\usepackage{amsmath}
\usepackage{fancyhdr}       
\usepackage{graphicx}       
\usepackage{cleveref}
\let\svthefootnote\thefootnote
\newcommand\freefootnote[1]{%
  \let\thefootnote\relax%
  \footnotetext{#1}%
  \let\thefootnote\svthefootnote%
}
\graphicspath{{media/}}     
\usepackage[format=plain,
            font=it]{caption}
\newcommand{\norm}[1]{\left\lVert#1\right\rVert}
\newcommand*{\horzbar}{\rule[.5ex]{2.5ex}{0.5pt}}

\usepackage{caption}
\usepackage{subcaption}

\title{Generative Modeling of Time-Dependent Densities via Optimal Transport and Projection Pursuit\\
}

\author{
Jonah Botvinick-Greenhouse$^*$ \\
  Center for Applied Mathematics\\
  Cornell University\\
  Ithaca, NY 14850, USA \\
  \texttt{jrb482@cornell.edu} \\
   \And
 Yunan Yang \\
 Department of Mathematics\\
  Cornell University\\
 Ithaca, NY 14853, USA \\
  \texttt{yunan.yang@cornell.edu} \\
  \And
 Romit Maulik$^*$ \\
 College of Information Science and Technology\\
 Pennsylvania State University\\
  University Park, PA 16802, USA\\
  \\
  Joint Appointment Faculty\\
  Mathematics and Computer Science Division\\
  Argonne National Laboratory\\
   Lemont, IL 60439, USA\\
  \texttt{rmaulik@anl.gov} \\
}

\begin{document}
\maketitle
\freefootnote{$^*$Corresponding authors}
\begin{abstract}
    Motivated by the computational difficulties incurred by popular deep learning algorithms for the generative modeling of temporal densities, we propose a cheap alternative which requires minimal hyperparameter tuning and scales favorably to high dimensional problems. In particular, we use a projection-based optimal transport solver [Meng et al., 2019] to join successive samples and subsequently use transport splines [Chewi et al., 2020] to interpolate the evolving density. When the sampling frequency is sufficiently high, the optimal maps are close to the identity and are thus computationally efficient to compute. Moreover, the training process is highly parallelizable as all optimal maps are independent and can thus be learned simultaneously. Finally, the approach is based solely on numerical linear algebra rather than minimizing a nonconvex objective function, allowing us to easily analyze and control the algorithm. We present several numerical experiments on both synthetic and real-world datasets to demonstrate the efficiency of our method. In particular, these experiments show that the proposed approach is highly competitive compared with state-of-the-art normalizing flows conditioned on time across a wide range of dimensionalities. 
\end{abstract}
\textbf{Deep learning algorithms for generative modeling of time-dependent data may face difficulties related to optimal hyperparameter selection and can be computationally expensive. In this work, we instead propose using a generative model based on numerical linear algebra and projection pursuit optimal transport, which requires only a few hyperparameter choices. We demonstrate the effectiveness of our model on synthetic data of stochastic dynamical systems and real-life examples of fish schooling and embryoid growth. These experiments show that our proposed approach is highly competitive with a state-of-the-art deep learning algorithm that learns to sample from time-dependent densities in terms of efficiency, accuracy, and interpretability.}

\section{Introduction}

\subsection{Motivation}

In recent times, there have been several studies that have investigated the use of machine learning algorithms for learning time-varying stochastic processes. Such processes are visible in many applications ranging from geoscience, bioscience, and engineering to computer vision. In particular, deep learning algorithms, such as neural network parameterized normalizing flows, neural ordinary differential equations, diffusion models, and generative adversarial networks, have shown remarkable advances in learning and enabling rapid sampling from these stochastic processes. Such advances are further pronounced for very high-dimensional systems where classical methods are seen to saturate their effectiveness. However, the effective use of deep learning is frequently hampered by difficulties associated with computational cost as well as optimal hyperparameter selection. In this article, we propose a novel approach based on projection-pursuit optimal transport, which learns to sample from the densities of time-varying stochastic processes. It is competitive (both in terms of computational cost and accuracy) with a state-of-the-art deep learning algorithm (given by the neural spline flow). Crucially, our proposed method requires few hyperparameter choices by the user in contrast with most neural network-based methodologies. Thus, our main contributions to this work are as follows:
\begin{enumerate}
    \item We implement a projection-pursuit optimal transport-based method to learn maps between time-varying densities from snapshots of particles sampled from these densities.
    \item We utilize a state-of-the-art cubic spline-based interpolation technique for trajectories of samples from the approximated densities, allowing us to smoothly interpolate between training data snapshots in time. 
    \item We thoroughly compare our approach with a state-of-the-art neural network parameterized normalizing flow, demonstrating competitive performance for several systems over a wide range of dimensions.
\end{enumerate}
Throughout, a "snapshot" refers to a collection of samples drawn from the underlying density at a fixed time. 
\subsection{Related Work}
Data-driven approaches for constructing models of complex dynamical systems have received considerable attention within several fields of study. Stochastic differential equations (SDEs) are often used to model physical processes whose dynamics are either probabilistic in nature or well-approximated by randomness. By leveraging prior knowledge about the functional form of a given SDE, one can use parameter estimation techniques to calibrate governing equations to available sample path data \cite{golightly2010markov,mbalawata2013parameter,NIELSEN200083,YANG2020110026}. Moreover, advancements in machine learning have led to models which parameterize the drift and diffusion of SDEs by neural networks \cite{jia2019neural,kidger2021neural, liu2019neural}. Other methods include the use of variational inference \cite{salimans2015markov}, which assumes a Gaussian structure for the target probability density function. However, such approaches rely on prior knowledge of the noise process that drives the dynamics and may be infeasible if this information is unavailable. 

Notably, \cite{both2019temporal,feng2021solving,Lu_2022,rasul2021multivariate,scholler2021flomo} overcome this challenge by deploying generative machine learning frameworks based on normalizing flows to learn the time-dependent density of trajectories without imposing any assumptions on a noise process or underlying functional form of the dynamics. A normalizing flow constructs a map from a known reference density to a target density via a sequence of parameterized invertible transformations \cite{Kobyzev_2021}. These transformations may also rely on the neural ordinary differential equation architecture \cite{NEURIPS2018_69386f6b} as the parameterization algorithm in which they are considered continuous normalizing flows~\cite{deng2021continuous}. One can also learn a time-dependent target density using the so-called conditional normalizing flows~\cite{winkler}. 

Other assumption-free methods for learning time-varying stochastic processes include using generative adversarial networks \cite{goodfellow2020generative}, which is a class of algorithms where two neural networks compete with each other in the form of a zero-sum game. A discriminator is trained to detect fake samples of the distribution, while a generator attempts to fool the discriminator. Given samples from a training distribution, this technique learns to generate new data with the same statistics as the training set. However, these methods are challenging to train and require significant inductive biases to adapt to scientific applications \cite{wiese2020quant,yang2020physics}. Newer techniques for learning densities include diffusion models inspired by non-equilibrium thermodynamics, which define a Markov chain of diffusion steps that adds random noise to training data and subsequently learn the inversion of this process, using a neural network to generate synthetic data~\cite{song2020score}. 

Training neural network-based models such as normalizing flows is compute-intensive, and this is due to the requirement of large-scale nonconvex optimizations as well as expensive hyperparameter and neural architecture searches. Therefore, it is desirable to investigate alternatives that can accelerate workflows reasonably by reducing the number of restarts for new hyperparameters and architecture choices. In the following, we put forth a competitive new approach.

\subsection{Our Approach}

\begin{figure}[h!]
    \centering
    \includegraphics[trim = {0cm 3.5cm 0cm 3.5cm},width= \textwidth]{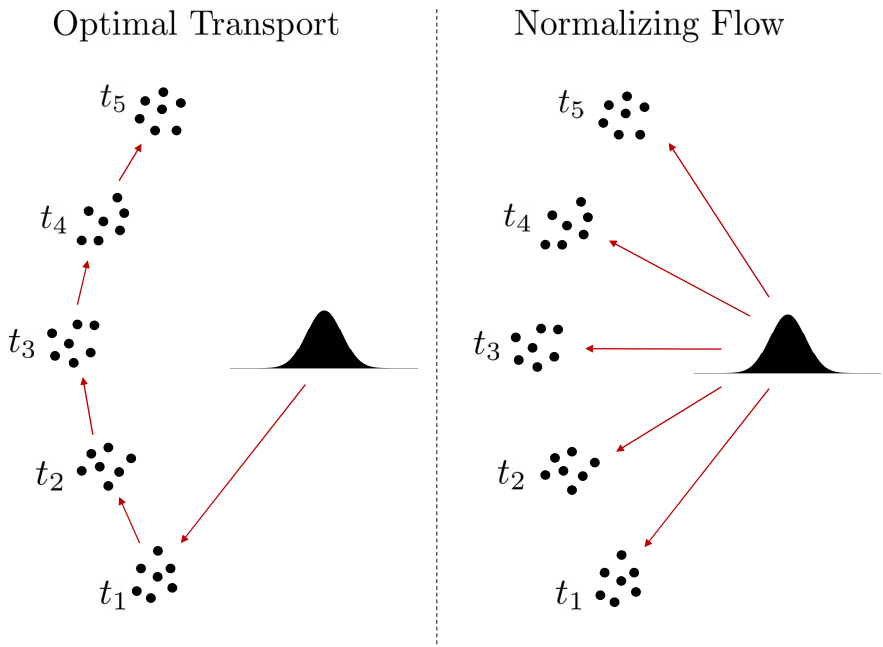}
    \caption{\textbf{Illustration of the two generative models.} The left panel shows a generative model obtained by mapping between successive snapshots via optimal transport maps, whereas the right panel shows the generative model obtained by mapping a known reference distribution to each snapshot.}
    \label{fig:F1}
\end{figure}

As mentioned above, normalizing flows often require large neural network architectures, which demand a long training time and careful hyperparameter tuning. Due to these challenges, we propose an alternative approach based upon optimal transport (OT)~\cite{villani2009optimal} and projection pursuit~\cite{friedman1981projection}, which can circumvent these difficulties. Rather than learning an invertible map between a reference density and several snapshots of samples, we learn an optimal transport map (OTM) from a reference density to the first snapshot of samples and join the remaining snapshots with successive optimal transport maps. This provides a generative model for the dynamics when samples were collected. With additional assumptions on the regularity of the stochastic process, the learned optimal transport maps can also be used to interpolate between snapshots smoothly. The contrast with a normalizing flow-based generative model is illustrated in Figure~\ref{fig:F1}. Modeling stochastic systems using optimal transport in this way has several computational benefits. Notably, if the inference data is sampled at a sufficiently high frequency, then the optimal transport maps will not deviate significantly from the identity and will therefore be cheap to approximate. Moreover, the procedure can be easily paralleled since the approximation of the optimal map between two snapshots is unaffected by the optimal maps at other times. While OT has been used to model time-dependent densities in~\cite{benamou2019second,https://doi.org/10.48550/arxiv.2010.12101} and has also been combined with normalizing flows~\cite{https://doi.org/10.48550/arxiv.2002.04461}, our approach investigates how it may be deployed for high-dimensional systems, representative of real-world applications, by using projection pursuit regression, thereby making it competitive with techniques such as the normalizing flow.

We accelerate the process of finding the OT map using the Projection Pursuit Monge Map (PPMM) algorithm presented in~\cite{https://doi.org/10.48550/arxiv.2106.05838}. The approach is inspired by projection pursuit regression~\cite{friedman1981projection}, as it seeks to estimate high-dimensional optimal transport maps by learning several one-dimensional optimal transport maps along the most "informative" projection directions. The PPMM algorithm is an iterative approach, and at the $k$-th step of the algorithm, the projection direction is selected to maximize the discrepancy (measured using the so-called sliced average variance estimation (SAVE)~\cite{doi:10.1080/03610920008832598}) between projections of the current samples and target samples. The PPMM algorithm empirically exhibits fast convergence and scales well to high-dimensional problems. It must be noted that the PPMM procedure used to approximate optimal transport maps is based solely on numerical linear algebra instead of minimizing a nonconvex objective function. It can be analyzed and controlled using standard tools in numerical analysis. 

After using PPMM to learn optimal transport maps between snapshots, we use the transport splines framework in~\cite{https://doi.org/10.48550/arxiv.2010.12101} to quickly interpolate the samples. While various methods for performing interpolation in Wasserstein space exist, e.g., \cite{benamou2019second}, we elect to use transport splines for their ease of implementation and computational feasibility. Indeed, the transport splines approach reduces the problem of interpolating a curve of measures to the Euclidean interpolation of Lagrangian trajectories, which are determined by the optimal coupling between snapshots. Through several numerical experiments on synthetic and real-world data sets, we illustrate that the proposed method exhibits faster convergence and higher accuracy in many situations than state-of-the-art neural spline flows~\cite{https://doi.org/10.48550/arxiv.1906.04032}. As the dimensionality of the state space becomes sufficiently large (e.g., $d = 30$), we observe that this performance gap becomes small, but that our proposed method is still more computationally efficient.

The rest of the paper is structured as follows. In \Cref{sec:background}, we review the necessary background on optimal transport, the PPMM algorithm, and normalizing flows. In~\Cref{sec:methods}, we discuss how we adapt both PPMM and normalizing flows to the time-dependent setting and review the generative algorithms shown in Figure~\ref{fig:F1} in detail. In~\Cref{sec:experiments}, we provide numerical results based on synthetic snapshot data from high-dimensional Fokker--Planck equations and real-world biological applications to fish schooling and embryoid growth. Conclusions follow in~\Cref{sec:conclusions}.

\section{Background} \label{sec:background}
This section provides an overview of optimal transport and normalizing flows as tools for constructing generative models of stationary densities. In \Cref{subsec:PPMM}, we review the theory of optimal transport and introduce the PPMM algorithm. In \Cref{subsec:NSF}, we provide an overview of normalizing flows as well as the neural spline flow architecture~\cite{https://doi.org/10.48550/arxiv.1906.04032}, which we use for our numerical experiments in \Cref{sec:experiments}. 

\subsection{Projection Pursuit Monge Map}\label{subsec:PPMM}
The Projection Pursuit Monge Map (PPMM) algorithm introduced in \cite{https://doi.org/10.48550/arxiv.2106.05838} provides an efficient approach to computing optimal transport maps in high dimensions. The method utilizes tools from projection pursuit regression \cite{friedman1981projection}, as well as the sufficient dimension reduction technique of Sliced Average Variance Estimation (SAVE) \cite{doi:10.1080/03610920008832598}. At each iteration, the SAVE procedure is used to choose a projection direction that maximizes the discrepancy between the variance of projections of the current and target samples. A one-dimensional OT map is then learned along this projection direction and is used to transport the collection of current samples accordingly. To discuss the PPMM algorithm more precisely, we now introduce a few concepts from the theory of optimal transport. 

Let $X:\mathbb{R}^d\to \mathbb{R}$ and $Y:\mathbb{R}^d\to \mathbb{R}$ be two continuous random variables, with corresponding densities $p_X$ and $p_Y$. We say that a measurable function $\phi: \mathbb{R}^d\to \mathbb{R}^d$ is a transport map between $p_X$ and $p_Y$ if for all Borel measurable sets $B$ it holds that $\mu_X(\phi^{-1}(B)) = \mu_Y(B)$, where $\mu_X$ and $\mu_Y$ denote the measures induced by the densities $p_X$ and $p_Y$, respectively. Since $\mu_X$ and $\mu_Y$ are absolutely continuous by assumption, we also have that $\phi$ is a transport map if $\phi$ is invertible and \eqref{eq:cov} holds:
\begin{equation}\label{eq:cov}
    p_{Y}(x) = p_{X}(\phi^{-1}(x)) | \det D \phi^{-1} (x)|.
\end{equation}
Though there may be many such transport maps, the goal is to recover one that is optimal in some sense.  Towards this, we introduce a function $c(x,y)$, which represents the cost of transporting a point mass $x\in \mathbb{R}^d$ to another point mass $y\in\mathbb{R}^d$. Given a transport map $\phi:\mathbb{R}^d\to \mathbb{R}^d$ and cost $c:\mathbb{R}^d\times \mathbb{R}^d\to \mathbb{R}$, we define the overall transport cost as 
\begin{equation}\label{eq:TC}
    \int_{\mathbb{R}^d}c(x,\phi(x)) p_X(x) dx.
\end{equation}
Typically, one fixes $p \in [1,\infty)$ and uses the cost $c(x,y) = |x-y|^p$, where $|\cdot|$ denotes the Euclidean norm. Provided that $X$ and $Y$ have finite moments of order $p$, this construction gives rise to the $p$-Wasserstein distance
\begin{equation}\label{E2}
    W_p(p_X,p_Y) := \Bigg( \inf_{\phi \in \Phi} \int_{\mathbb{R}^d} |x - \phi(x)|^p p_X(x) dx\Bigg)^{1/p}.
\end{equation}
In \eqref{E2}, $\Phi$ denotes the collection of all transport maps from $p_X$ to $p_Y$. For a particular choice of $p\in [1,\infty)$, we will hereafter denote by $\phi_*\in \Phi$ the transport map for which the infimum \eqref{E2} is achieved. 

The optimal transport map $\phi_*$ and $p$-Wasserstein distance are popular tools in the machine learning community for training generative models. While \eqref{E2} can be used as the loss function to train a given model~\cite{arjovsky2017wasserstein}, one can also use the mapping $\phi_*$ as the generative model itself \cite{rout2021generative}. We adopt the latter approach. Given $N$ samples $\mathbf{X},\mathbf{Y}\in \mathbb{R}^{N\times d}$
drawn from the densities $p_X$ and $p_Y$ respectively, 
the PPMM algorithm constructs $k$-step approximations $\widehat{\phi}_k$ (see~\Cref{alg:PPMM}) and $\widehat{W}^{(k)}_2$ (see~\eqref{eq:W2app}) of the true OT-map $\phi_*$ and the $2$-Wasserstein distance~\eqref{E2}, respectively. 

\begin{algorithm}[H]
\SetAlgoLined
\textbf{Input: }  $\mathbf{X},\mathbf{Y}\in\mathbb{R}^{N\times d}$. \\
	Set $\mathbf{X}_0 = \mathbf{X}$  and
	$k = 0$.\\
	\textbf{Until convergence:}\\
\qquad	(a) Use SAVE to compute the projection direction $p_k\in\mathbb{R}^d$ between $\mathbf{X}_k$ and $\mathbf{Y}$.\\
\qquad	 (b) Compute the one-dimensional OT-map $\eta_k$ between the projections $\mathbf{X}_kp_k$ and $\mathbf{Y}p_k$.\\
\qquad	(c) Set $\mathbf{X}_{k+1} = \mathbf{X}_{k}+\big(\eta_k(\mathbf{X}_kp_k)-\mathbf{X}_kp_k\big)p_k^T$ and $k = k+1$.\\
\textbf{Output: } $\{p_{\ell}\}_{\ell=0}^{k-1}$ and $\{\eta_{\ell}\}_{\ell=0}^{k-1},$
which form the OT approximation $\widehat{\phi}_k:\mathbb{R}^{ d}\to \mathbb{R}^{d}$ given in~\eqref{eq:PPMM_eq}.
\caption{Projection Pursuit Monge Map (PPMM) \cite{https://doi.org/10.48550/arxiv.2106.05838}\label{alg:PPMM}}
\end{algorithm}

We remark that the output $\widehat{\phi}_k$ of the PPMM algorithm is defined recursively by setting $\widehat{\phi}_0(x) = x$ and
\begin{equation}\label{eq:PPMM_eq}
\widehat{\phi}_{\ell+1}(x) = \widehat{\phi}_{\ell}(x) + \Big(\eta_{\ell}\Big( \widehat{\phi}_{\ell}(x) p_{\ell}\Big) - \widehat{\phi}_{\ell} (x) p_{\ell}\Big)p_{\ell}^T, \qquad 0 \leq \ell \leq k-1 . 
\end{equation}
In step (a) of Algorithm~\ref{alg:PPMM}, the projection direction $p_k$ is chosen to maximize the discrepancy between $\text{Var}(\mathbf{X}^{(k)} p_k)$ and $\text{Var}(\mathbf{Y} p_k)$. In step (b), a one-dimensional (1D) OT map $\eta_k$ is computed between the projected samples $\mathbf{X}_k p_k$ and $\mathbf{Y}p_k$. Finally, in step (c), the collection $\mathbf{X}_k$ of current samples is updated according to the transport map $\eta_k$. We note that in~\cite{https://doi.org/10.48550/arxiv.2106.05838}, the 1D OT maps $\{\eta_{\ell}\}_{\ell=0}^{k-1}$ from (b) are computed via sorting and interpolation. This is because the 1D optimal transport problem has an analytical solution, which the Quicksort algorithm can efficiently compute in $\mathcal{O}(N\log N)$ time. In Section \ref{subsubsec:reg}, we propose an alternative approach for computing the 1D maps $\eta_{\ell}$, which is more efficient when $N$ is large, acts as a helpful regularization when $N$ is small, and extends the domain of $\eta_{\ell}$ to all of $\mathbb{R}$. By modeling each $\eta_{\ell}$ as a mapping from $\mathbb{R}\to \mathbb{R}$, we then have that the approximation $\widehat{\phi}_k$ is a mapping  from $\mathbb{R}^d \to \mathbb{R}^d$, as given in \eqref{eq:PPMM_eq}. This perspective motivates the use of PPMM as a generative model. We remark that weak convergence of \Cref{alg:PPMM} has been proven (see \cite[Theorem 3]{https://doi.org/10.48550/arxiv.2106.05838}) and that its computational complexity has been studied (see \cite[Section 2]{https://doi.org/10.48550/arxiv.2106.05838}).

In our implementation of PPMM in \Cref{sec:experiments}, we also enforce a relative stopping criterion which terminates the algorithm after convergence of the approximate 2-Wasserstein distance,
\begin{equation}\label{eq:W2app}
    \widehat{W}_2^{(k)}( \mathbf{X},\mathbf{Y}) := \Bigg( \frac{1}{N}\sum_{i=1}^N\norm{\widehat{\phi}_k(x_i) - x_i}^2\Bigg)^{1/2},
\end{equation}
begins to slow down. That is, we specify a tolerance $\alpha \in [0,1]$ and terminate PPMM when 
\begin{equation}\label{eq:conv}
   \frac{ \big|\widehat{W}_2^{(k(\alpha))}(\mathbf{X},\mathbf{Y}) - \widehat{W}_2^{(k(\alpha)-1)}(\mathbf{X},\mathbf{Y})\big|}{\big|\widehat{W}_2^{(k(\alpha))}(\mathbf{X},\mathbf{Y})\big|} \leq \alpha,
\end{equation}
where $k(\alpha)\in \mathbb{Z}_+$ is the  the last iteration number. 
Note that the convergence criteria \eqref{eq:conv} does not rely on a knowledge of $W_2(p_X,p_Y)$. Thus, we can use it in applications where the ground truth solution is unknown. Moreover, since we will ultimately be interested in applying PPMM to many pairs of snapshots simultaneously, using the convergence criteria~\eqref{eq:conv} ensures that we spend most of our computational efforts learning the optimal transport maps along the dominant directions. While we have stated the PPMM algorithm for the case when $\mathbf{X},\mathbf{Y} \in \mathbb{R}^{N\times d}$, we note that it can also be used to learn the OTM between distributions when the sample sizes are not equal. 

\subsection{Regularizing the Optimal Transport Maps}\label{subsubsec:reg}
The one-dimensional optimal maps $\eta_{\ell}$ in Algorithm \ref{alg:PPMM} are computed according to the quicksort algorithm, which runs in $\mathcal{O}( N \log N)$ time. However, when $N$ is sufficiently large, it can be computationally advantageous to instead form histogram approximations using $B \ll N$ cells of the underlying samples and learn the optimal map according to the weighted histogram cell centers. More specifically, assume we want to learn the optimal map between densities $\rho_1,\rho_2:\mathbb{R}\to [0,\infty)$ from the observed samples  $\mathbf{x},\mathbf{y}\in \mathbb{R}^N$ with $\mathbf{x}\sim \rho_1$ and $\mathbf{y}\sim \rho_2$. The true optimal map is given by $G^{-1} \circ F$, where $F$ and $G$ denote the cumulative distribution functions (CDFs) of $\rho_1$ and $\rho_2$, respectively \cite{peyre2019computational}. Thus, we use a  kernel density estimate (KDE) based upon the Gaussian kernel to obtain approximations $\hat{F}$ and $\hat{G}$ of the CDFs and compute the underlying optimal map as
\begin{equation}\label{eq:optimal1d}
   \eta(x):=\begin{cases}  f_2(f_1(x)) & x\in \Upsilon\\
    \text{Id}(x) & x\not\in \Upsilon
    \end{cases}  , \qquad  f_1:= \texttt{Interp}(\mathbf{z},\hat{F}), \qquad f_2:=\texttt{Interp}(\hat{G},\mathbf{z}).
\end{equation}
In \eqref{eq:optimal1d}, $\texttt{Interp}(\cdot,\cdot)$ denotes a piecewise linear interpolation and $\mathbf{z}\in \mathbb{R}^B$ is a vector containing $B$ equally spaced histogram cell centers over the interval $\Upsilon: = [ \min\{\mathbf{x},\mathbf{y}\} - L, \max\{\mathbf{x},\mathbf{y}\}+L]\subseteq \mathbb{R}$. The constant $L > 0$ should be chosen large enough such that the effective support of the approximate densities is strictly contained in $\Upsilon$. We remark that the KDE of the binned samples can be evaluated at the equidistant grid points $\mathbf{z}\in \mathbb{R}^B$ in $\mathcal{O}(B\log B)$ time based on an approach using the fast Fourier transform (FFT) \cite{tommy_odland_2018_2392268}. Moreover, to enforce strict monotonicity of the approximate CDFs, $\hat{F}$ and $\hat{G}$, we rescale the densities obtained from the KDE by adding a small positive constant $\varepsilon >0 $ and renormalizing. 

The density estimation can heavily depend on the KDE bandwidth, so the bandwidth should be chosen carefully and in accordance with the observed data. A bandwidth that is too small will overfit the training data, whereas a bandwidth that is too large may result in an over-smoothed distribution.
While the use of an FFT-based KDE can be computationally advantageous when modeling the OT maps $\eta_{\ell}$ for large values of $N$, it can also serve as a useful regularization in the low-data limit. Indeed, when $N$ is small, the selection of a sufficiently large bandwidth for the KDE effectively injects additional noise in the under-representative training samples. This helps compensate for the small number of observations and prevents overfitting. Otherwise, we select the bandwidth according to either Scott's method \cite{scott2012multivariate} when the data is close to unimodal, or the improved Sheather--Jones (ISJ) method \cite{10.1214/10-AOS799} when the data is multimodal. 

\subsection{Neural Spline Flows}\label{subsec:NSF}
Normalizing flows deliver alternative approaches to the generative modeling of samples drawn from a probability distribution. While the PPMM algorithm only provides a generative model for samples of the underlying density, a normalizing flow learns the density itself. Moreover, while OT-based methods seek to construct an optimal mapping between samples between reference and target densities, normalizing flows learn a transport map that is not necessarily optimal in the sense of~\eqref{E2}. 

We now describe the optimization procedure by which a normalizing flow learns a transport map from a known reference density to a target density. 
Specifically, assume that $p_X: \mathbb{R}^d \to \mathbb{R}$ is the probability density function of the continuous random variable $X:\mathbb{R}^d\to \mathbb{R}$ and that $g:\mathbb{R}^d\to \mathbb{R}^d$ is invertible with $f:=g^{-1}.$ For the random variable $Y:= g\circ X$ with corresponding density $p_Y$, the change of variables formula for probability densities is given in~\eqref{eq:cov} with the map $\phi$ replaced by $g$ here. By rewriting $g = g_K \circ \dots \circ g_1$ as the composition of $K$ bijective functions, the determinant $\det D f$ can be evaluated as
$$\det D f(x) = \prod_{i=1}^K \det D f_i(x_i), \qquad x_i := \begin{cases}
x & i = K\\
(f_{i+1}\circ \dots \circ f_K)(x) & 1\leq i < K 
\end{cases}, \qquad f_i := g_i^{-1}.
$$
If one samples the data matrix $\mathbf{Y}\in \mathbb{R}^{N\times d}$ from an underlying density $p_Y$, a normalizing flow can be used to estimate $p_Y$ by parameterizing the family of bijections $g = g_{\theta}$ by $\theta \in \Theta$ and performing the optimization
\begin{equation}\label{eq:MLE}
    \max_{\theta\in \Theta}\mathcal{L}_{\theta},\hspace{1cm} \mathcal{L}_{\theta}:= \frac{1}{N}\sum_{n=1}^N \Big( \log p_X ( f_{\theta}(\mathbf{Y}_n)) + \log \big| \det Df_{\theta}(\mathbf{Y}_n)\big|\Big).
\end{equation}
In \eqref{eq:MLE} we have written $\mathbf{Y}_n$ to denote the $n$-th sample of $\mathbf{Y}$, and again $f_{\theta}:=g_{\theta}^{-1}.$ The density $p_X$ is a known reference distribution (usually taken to be Gaussian). During training, the flow learns to transport the samples $\mathbf{Y}$ to samples of $p_X$ via the mapping $f_{\theta}.$ Since $f_{\theta}$ is bijective with an inverse $g_{\theta},$ the flow can also be used to generate new samples from $p_Y$ once it has been trained, simply by applying the mapping $g_{\theta}$ to new samples drawn from $p_X$. 

We use neural spline flows~\cite{https://doi.org/10.48550/arxiv.1906.04032} for their state-of-the-art capabilities in modeling high-dimensional densities. Specifically, we study neural spline flows based upon coupling transforms functions $\psi_m:\mathbb{R}^d\to \mathbb{R}^d$ of the form 
\begin{equation}\label{eq:couple}
    \psi_m(x_1,\dots ,x_{m-1}, x_m, \dots, x_d)= (x_1,\dots,x_{m-1}, h_{\theta_m}(x_m),\dots, h_{\theta_d}(x_d)),
\end{equation}
where the family $\{h_{\theta_j}\}_{j=m}^d$ of functions are invertible and parameterized by $(\theta_m,\dots,\theta_d)$, which is set as the output of a neural network with inputs $\{x_1,\ldots, x_{m-1}\}$, i.e., $\text{NN}(x_1,\dots,x_{m-1})$. Essentially, \eqref{eq:couple} splits an input vector at the index $1\leq m\leq d$ and applies a transformation paramaterized by  $(x_1,\dots,x_{m-1})$ to the entries $(x_m,\dots, x_d).$ It is also common to permute the entries of the input $x \in \mathbb{R}^d$ before passing through the transform \eqref{eq:couple}. Coupling transforms enable a seamless computation of the Jacobian determinant, as  $D\psi_m$ is by construction block triangular. Indeed, the Jacobian determinant can be evaluated as the product 
$$\det D \psi_m = \prod_{j=m}^d \frac{\partial h_{\theta_j}}{\partial x_j}.$$

To improve the approximation power of coupling transforms \eqref{eq:couple}, neural spline flows parameterize the functions $h_{\theta_j}$ as families of monotonic rational-quadratic splines. To allow all parameters of the flow to interact with one another, the flow architecture typically alternates between spline-based coupling layers and linear layers. The linear layers are based upon the $PLU$ decomposition. where at the beginning of training, the permutation matrix $P$ is fixed and the matrices $L$ and $U$ are learned. As in \cite{https://doi.org/10.48550/arxiv.1906.04032}, a single step of the flow consists of a spline layer and a 
linear layer. 

Following \cite{https://doi.org/10.48550/arxiv.1906.04032}, we refer to the number of knots in the monotonic rational-quadratic splines $h_{\theta_j}$ as the number of bins. A tail bound of $b > 0$ is prescribed for the splines $h_{\theta_j}$ such that the functions $h_{\theta_j}(x)$ are extended to be strictly linear when $|x| > b$. This allows the flow to evaluate unconstrained inputs; see \cite[Fig. 1]{https://doi.org/10.48550/arxiv.1906.04032}. Moreover, the neural network which computes the parameters $\theta_j$ in the coupling transform is taken as a residual network with pre-activation blocks. Our experiments in \Cref{sec:experiments} utilize ten-step RQ-NSFs with $8$ bins, $2$ residual blocks with $128$ hidden features, and a tail bound of $b=3$. This is the same architecture that achieved state-of-the-art density estimation results in \cite{https://doi.org/10.48550/arxiv.1906.04032}. Since we are interested in efficient density estimation, we also test a Reduced RQ-NSF architecture, which consists of only two steps. This architecture can be trained quickly and provides an expedient means for calculating likelihoods. However, it reduces the approximation power of the flow compared to the ten-step framework.

\section{Methods}\label{sec:methods}
In this section, we formally introduce Dynamic PPMM and adapt the RQ-NSF framework to the time-dependent setting. We begin in \Cref{subsec:DPPMM} by
discussing an alternative approach to computing the one-dimensional OT maps in the PPMM algorithm. We then proceed by formally stating the Dynamic PPMM algorithm, reviewing the transport splines algorithm presented in~\cite{https://doi.org/10.48550/arxiv.2010.12101}, and commenting on the computational complexity of training the model. We conclude in \Cref{subsec:DNSF} by explaining how conditional normalizing flows \cite{nflows} can be used to adapt the neural spline flow framework from \Cref{subsec:NSF} to the time-dependent setting. 
\subsection{Dynamic PPMM}\label{subsec:DPPMM}
We will assume that $\rho(x,t)$ is a time-dependent density on $\mathbb{R}^d$  and that at each observation time $t_1 < t_2 < \dots < t_M$ we draw $N_j$ samples $\mathbf{X}^{(t_j)}\in \mathbb{R}^{N_j\times d}$ from $\rho(x,t_j)$. We now aim to use our collection $\{\mathbf{X}^{(t_j)}\}_{j=1}^M$ of time-dependent samples to sample from $\rho(x,t)$, for any $t\in [t_1,t_M].$ Our proposed approach involves using the PPMM algorithm to join successive snapshots via optimal transport maps, and subsequently applying a Euclidean interpolation algorithm to individual trajectories between the snapshots.
\subsubsection{The Algorithm}\label{subsubsubsec:algo}

We now state the Dynamic PPMM algorithm, which for convenience, will be frequently referred to as D-PPMM. 
\begin{algorithm}[H]
	\SetAlgoLined
	\textbf{Input: } Measurement times $\{t_j\}_{j=1}^M$ and samples $\{\mathbf{X}^{(t_j)}\}_{j=1}^{M
}$.\\
	\textbf{Training:}\\
\qquad 	Sample the rows of $\mathbf{X}^{(t_0)} \in \mathbb{R}^{N_1\times d}$ i.i.d. from $ \mathcal{N}(\nu,\Gamma)$. \\
\qquad \textbf{For} $j \in \{0,\dots,M-1\}$  \textbf{do}:\\
\qquad \qquad Use PPMM (\Cref{alg:PPMM}) to estimate the OTM $\widehat{\phi}^{(t_j)}_{k_j(\alpha)}$ between $\mathbf{X}^{(t_j)}$ and $\mathbf{X}^{(t_{j+1})},$\\
\qquad \qquad where each 1D OTM is given by the procedure in \Cref{subsubsec:reg}.\\
\textbf{Testing:}\\
\qquad Sample the rows of $\mathbf{Y}^{(t_0)} \in \mathbb{R}^{\tilde{N}\times d} $ i.i.d. from $\mathcal{N}(\nu,\Gamma)$.\\
\qquad \textbf{For} $j\in \{1,\dots, M\}$ \textbf{do}:\\
\qquad \qquad Assign $\mathbf{Y}^{(t_j)} = \widehat{\phi}^{(t_{j-1})}_{k_j(\alpha)}(\mathbf{Y}^{(t_{j-1})})$\\
	\textbf{Output:} New samples $\{\mathbf{Y}^{(t_j)}\}_{j=1}^M$. 
	\caption{Dynamic PPMM (D-PPMM)\label{alg:D-PPMM}}
\end{algorithm}

Note that $\mathbf{Y}^{(t_0)}$ contains $\tilde{N}$ samples, where $\tilde{N}$ is the number of new samples we wish to generate using D-PPMM. The D-PPMM algorithm has only a handful of tunable parameters, most of which are related to the construction of the one-dimensional optimal maps; see \eqref{eq:optimal1d}. The hyperparameters that can affect the performance of Algorithm~\ref{alg:D-PPMM} include the convergence tolerance $\alpha \in [0,1]$, the KDE bandwidth selection method, the number of histogram cells $B$, the base normal distribution $\mathcal{N}(\nu,\Gamma)$, and the constants $L,\varepsilon > 0$; see \Cref{subsubsec:reg}. We find that the algorithm's performance is relatively insensitive to the choice of $\nu$, $\Gamma$, $L$, and $\varepsilon$. Thus, throughout our experiments in \Cref{sec:experiments}, we leave the base normal distribution fixed with $\nu = 0$ and $\Gamma = 0.01 I$ where $I$ is the identity matrix, and we set $\varepsilon = 10^{-8}$ and $L = 0.1$ or $L = 0.25$.

\subsubsection{Interpolation via Transport Splines}\label{subsubsec:ts}
To extend Algorithm \ref{alg:D-PPMM} to a generative model for $\rho(x,t)$, for any $t\in [t_1, t_M]$, we use the transport splines algorithm introduced in \cite{https://doi.org/10.48550/arxiv.2010.12101}. This approach involves first coupling the snapshots via optimal transport maps, e.g., computing the output $\smash[b]{\{\mathbf{Y}^{(t_j)})\}_{j=1}^M}$ of~\Cref{alg:D-PPMM}. Then, for each $n\in \{1,\dots, \tilde{N}\}$, the trajectories $\smash[b]{\{(t_j,\mathbf{Y}_n^{(t_j)})\}_{j=1}^M}$ are interpolated in Euclidean space via cubic splines. That is, for each fixed $n$,  we compute its spline interpolant $p_n: [t_1,t_M]\to \mathbb{R}^d$:
\begin{equation}\label{eq:spline}
   p_n := \texttt{Spline}\Big([t_1, t_2,\dots, t_M], [\mathbf{Y}_n^{(t_1)},\mathbf{Y}_n^{(t_2)},\dots, \mathbf{Y}_n^{(t_M)}]\Big) ,
\end{equation}
where $\texttt{Spline}(\cdot,\cdot)$ denotes a cubic B-spline interpolation. Then, for arbitrary $t\in [t_1,t_M]$, we form the interpolating snapshot $\mathbf{Y}^{(t)}$ by evaluating
\begin{equation}
\mathbf{Y}^{(t)} =\left[
  \begin{array}{ccc}
    \horzbar & p_{1}(t) & \horzbar \\
    \horzbar & p_{2}(t) & \horzbar \\
             & \vdots    &          \\
    \horzbar & p_{\tilde{N}}(t) & \horzbar
  \end{array}\right] \in \mathbb{R}^{\tilde{N} \times d}.
\end{equation}
We remark that no regularity assumptions are required if we simply aim to construct a generative model at the measurement times $t_1<t_2<\dots < t_M$, as in Algorithm \ref{alg:D-PPMM}. However, additional knowledge of the underlying stochastic process may be required to justify the use of cubic spline interpolation between measurement times, as the approach \cite{https://doi.org/10.48550/arxiv.2010.12101} was devised for stochastic processes with $C^2$ sample paths. Thus, the use of transport spline interpolation is not expected to be effective for discontinuous stochastic processes, such as Cauchy noise or a jump process. Therefore, our application of transport splines to real data in \Cref{subsec:fish} and \Cref{subsec:embryo} assumes sufficient regularity of the underlying processes. The transport splines algorithm has theoretical approximation guarantees (see \cite[Theorem 2]{{https://doi.org/10.48550/arxiv.2010.12101}}) and is computationally expedient, as it reformulates the problem of interpolating measures into the Euclidean interpolation of individual trajectories. As the time between observations decreases, so does the approximation power of the transport spline interpolation.

\subsubsection{Computational Complexity}
We now discuss the computational requirements for training Algorithm \ref{alg:D-PPMM}. For simplicity, we will assume that the Scott bandwidth selection rule is used, and we will write $k_j(\alpha) = k_j$ as the number of PPMM iterations which are performed at each timestep. Using PPMM to estimate each optimal map $\smash[b]{\hat{\phi}_{k_j}^{(t_j)}}$ incurs a cost of $\mathcal{O}(k_j N_j d^2 + k_j B\log B)$; see~\cite{https://doi.org/10.48550/arxiv.2106.05838} and Section \ref{subsubsec:reg}. We also remark that empirical evidence suggests that $k_j = \mathcal{O}(d)$ is expected to yield good convergence of the PPMM algorithm; see \cite[Figure 8]{https://doi.org/10.48550/arxiv.2106.05838}. The first term in the complexity bound is due to the computation of the most informative projection direction via the SAVE algorithm, whereas the second term is the cost associated with learning a one-dimensional optimal map via the procedure outlined in \Cref{subsubsec:reg}. Thus, if we let $K = \max_{0\leq j \leq M-1 } k_j$ and $N = \max_{0\leq j \leq M-1 } N_j$, we have that the total cost of training D-PPMM is given by $\mathcal{O}(K N d^2 + K B\log B).$ Note that in this analysis, the cost associated with pushing the $N_j$ samples through the OT maps at each step of training, as well as the KDE bandwidth selection are absorbed in the $\mathcal{O}$-notation.

\subsection{Time-Conditioned RQ-NSF}\label{subsec:DNSF}
We now discuss how the normalizing flows from \Cref{subsec:NSF} may be adapted to the time-dependent setting. As in \Cref{subsec:DPPMM} we consider the problem of learning the temporal density $\rho(x,t)$ when $t\in [t_1,t_M]$, given the observations $\{\mathbf{X}^{(t_j)}\}_{j=1}^M$ at the times $\{t_j\}_{j=1}^M$. Similar to \cite{Lu_2022}, we accomplish this by viewing the underlying density as a distribution conditioned upon the time $t$, which we then model via a conditional normalizing flow~\cite{winkler}. In analogy with~\eqref{eq:cov}, we now seek a mapping $f:\mathbb{R}^d \times [t_1,t_M] \to \mathbb{R}^d$ such that  
\begin{equation}\label{eq:cov2}
    \rho(x,t) = p_{X\mid t}(f(x,t))|\det D_x f(x,t)|, \qquad x\in \mathbb{R}^d, \qquad t\in [t_1,t_M], 
\end{equation}
where $f(\cdot,t):\mathbb{R}^d\to \mathbb{R}^d$ is a bijection for each fixed $t\in [t_1,t_M]$. Note that in \eqref{eq:cov2} the base distribution $p_X = p_{X\mid t}$ is now conditioned on time, as well. Our goal is to learn a suitable base distribution $p_{X\mid t}$ and bijection $f(x,t)$ from the data $\smash[b]{\{\mathbf{X}^{(t_j)}\}_{j=1}^M}$ such that \eqref{eq:cov2} holds.  Towards this, we use \texttt{nflows} \cite{nflows} to simultaneously train an RQ-NSF (see \Cref{subsec:NSF}) along with two additional feed-forward networks, $\smash[b]{\text{NN}^{(1)}:\mathbb{R}\to \mathbb{R}}$ and $\smash[b]{\text{NN}^{(2)}:\mathbb{R}\to \mathbb{R}^{2d}}$, which learn the time-conditioning in the bijection $f(x,t)$ and the base distribution $p_{X\mid t}$, respectively. More specifically, the function $f(x,t)$ is modeled as an RQ-NSF with an additional dimension concatenated to account for the sampling time, the inputs of which are given by $\smash[b]{\text{NN}^{(1)}:\mathbb{R}\to \mathbb{R}}.$ Moreover, we parameterize $p_{X\mid t}$ as a conditional diagonal normal distribution $\mathcal{N}(\nu(t),\text{diag}(\sigma(t)))$ and model the functions $\nu:\mathbb{R}\to \mathbb{R}^d$ and $\sigma: \mathbb{R}\to \mathbb{R}^d$ using $\smash[b]{\text{NN}^{(2)}: \mathbb{R}\to \mathbb{R}^{2d}}.$ 

 We use mini-batch training in which each step of the optimization includes exactly the same number of samples from each observation time $t_j$. 
 We also use the Adam optimizer and decay the learning rate from $5\cdot 10^{-4}$ to 0 on a cosine schedule every five epochs. Moreover, we parameterize $\text{NN}^{(1)}$ as a four layer fully connected ReLU network with 100 nodes in each layer, and we parameterize $\text{NN}^{(2)}$ as a fully connected ReLU network with a single layer of 500 nodes. We note that the network architecture of both the time-conditioned Reduced RQ-NSF and time-conditioned standard RQ-NSF have hundreds of thousands of tunable parameters. There are clearly also many user-specified hyperparameters that can affect the performance of the time-conditioned normalizing flow. We have tried to make our conditional RQ-NSFs competitive through a manual hyperparameter search. In this regard, a clear benefit of the D-PPMM algorithm is that it has only a select few hyperparameters which can be tuned; see \Cref{subsubsubsec:algo}. 

To interpolate the density $\rho(x,t)$ between two measurement times $\{t_j,t_{j+1}\}$ using the conditioned normalizing flow, we simply evaluate the right hand side of \eqref{eq:cov2} for $t\in [t_j,t_{j+1}]$. Note that the interpolation technique here differs from \Cref{subsec:DPPMM}, where we instead utilized the optimal coupling between snapshots to interpolate individual trajectories. 

\section{Numerical Results}\label{sec:experiments}
In this section, we compare D-PPMM with a time-conditioned RQ-NSF through several numerical experiments\footnote{Our implementations of D-PPMM and the time-conditioned RQ-NSF are publicly available at \url{https://github.com/jrbotvinick/Dynamic-PPMM}.}. All experiments were conducted using an Intel i7-1165G7 CPU. In \Cref{subsec:dataprep}, we introduce the synthetic dynamical system examples and explain the data generation process. In \Cref{subsec:lowdim} we present results comparing the two methods in the low-dimensional setting when $d < 5$. In \Cref{subsec:highdim}, we move onto the high-dimensional setting and study convergence behavior of the generative models when $d \in \{10,20,30\}$. Finally, in \Cref{subsec:fish,subsec:embryo}, we study the interpolation capabilities of D-PPMM applied to two real-world datasets from biological applications and compare the results, both in terms of speed and accuracy, with the interpolation of a conditioned RQ-NSF.

\subsection{Data Preparation \& Experimental Setup}\label{subsec:dataprep}
 Throughout all experiments, we rescale the sample data to belong to the cube $[-1,1]^d \subseteq \mathbb{R}^d$ via an affine transformation. We also rescale the trajectory time interval to $[0,1]\subseteq \mathbb{R}$.  We now discuss the preparation of the data used in Sections \ref{subsec:lowdim} and \ref{subsec:highdim}.  Throughout, we seek generative models for the sample paths of stochastic differential equations of the form \begin{equation}\label{eq:SDE}
     dX_t = v(X_t;\theta) dt + \sqrt{2D} dW_t,
 \end{equation} where $v(\cdot;\theta)$ denotes a parameter-dependent velocity, $D>0$ is the  diffusion which for simplicity we assume to be constant, and $W_t$ denotes a Brownian motion. We remark that the sample paths \eqref{eq:SDE} are described by the probability flow of a Fokker--Planck equation, i.e., a density $\rho(x,t)$ governed by the partial differential equation (PDE)
 \begin{equation}\label{eq:FPE}
     \frac{\partial}{\partial t}\rho(x,t) + \nabla \cdot (\rho(x,t) v(x;\theta)) = D \Delta \rho(x,t).
 \end{equation}
 
 We will denote by $N$ the number of sample paths of \eqref{eq:SDE} which evolve over the time interval $[0,T]$. To obtain a numerical approximation of \eqref{eq:SDE}, we use the Euler--Maruyama discretization \cite[Section 5.2]{SP}, which assigns 
\begin{equation}\label{eq:EM}
   \mathbf{X}_n^{j+1} =\mathbf{X}_n^{j} + v(\mathbf{X}_n^{j})\Delta t + \xi_j\sqrt{2 D \Delta t},
\end{equation}
where $\xi_j \sim \mathcal{N}(0,I)$ are i.i.d from the standard $d$-dimensional normal distribution and $\Delta t > 0$. 
\begin{table}[h!]
\centering
\setlength\tabcolsep{4pt}
\begin{tabular}{|c|c|c|c|c|c|}
\hline
System & $v$ & $D$ & $\theta$ &  $\mathbf{X}^{(t_0)}$ & $T$ \\
\hline 
Van der Pol & $\begin{aligned}
  \dot{x}_1 &= x_2  \\ \dot{x}_2 &=  c(1-x_1^2)x_2-x_1
\end{aligned}$ & $2.5\cdot 10^{-3}$ & $c = 1$ & $\begin{aligned}
&\mathcal{N}(\nu, \sigma^2 I)\\
&\nu = (1,1)\\
&\sigma = 5\cdot 10^{-2} 
\end{aligned}$ & $6$\\ \hline
OU-Process & $\begin{aligned}\dot{x}_i &=-\lambda x_i\\
1&\leq i \leq d,\quad d\geq 2
\end{aligned} $ & $5\cdot 10^{-2}$ & $\lambda = 10^{-1}$ & $\begin{aligned}
& \frac{1}{2} \left(\mathcal{N}(\nu_1, \sigma^2 I)+\mathcal{N}(\nu_2, \sigma^2 I) \right)\\
&\nu_i = (10(-1)^i, 10,\dots, 10)\\
&\sigma = 5 \cdot 10^{-2} 
\end{aligned}$ & $15$\\
\hline

Lorenz-96 & $\begin{aligned}
   \dot{x}_i &= (x_{i+1} - x_{i-2})x_{i-1}-x_i +F\\
   x_{-1} &= x_{d-1}, \hspace{.1cm}
   x_0=x_d, \hspace{.1cm}
   x_{d+1} = x_1 \\
    1&\leq i \leq d,\quad d \geq 4 \end{aligned}$ & $5\cdot  10^{-3}$ & $F = 2$ & $\begin{aligned}
&\mathcal{N}(\nu, \sigma^2 I)\\
&\nu = (4,0,\dots,0)\\
&\sigma = 10^{-1} 
\end{aligned}$ & $\leq 5$\\
      \hline

   \end{tabular}
   
   \vspace{.25cm}
\caption{ Equations and parameters for the synthetic dynamical systems studied throughout this section.}\label{tabel1}
\end{table}

Throughout our experiments in \Cref{subsec:lowdim} and \Cref{subsec:highdim}, we study the Van der Pol oscillator, an Ornstein--Uhlenbeck process (OU-process), and the Lorenz-96 system with additive stochastic forcing. While the Van der Pol oscillator has a fixed dimensionality of $d = 2$, we can manually adjust the dimensionality of the OU-process and Lorenz-96 system. Table \ref{tabel1} details the specific parameter choices, diffusion coefficients, and initializations that we use throughout our experiments. After approximating the sample paths \eqref{eq:SDE} via \eqref{eq:EM} over the time interval $[0,T]$, we select $M$ snapshots $\smash[b]{\{\mathbf{X}^{(t_j)}\}_{j=1}^M}$ evenly spaced in time as the inference data during our experiments. Each experiment in Sections~\ref{subsec:lowdim}~and~\ref{subsec:highdim} uses a total of $M = 11$ snapshots and $N = 10^4$ samples per snapshot.

We now discuss the process by which we quantify the error of the two generative models.   Both models are trained using the same inference samples, while error is quantified by comparing new samples from the generative models with a separate held-out testing set from the ground truth dynamical systems. The error in the generated samples from a given model is then determined using the Maximum Mean Discrepancy (MMD). Given collections of samples $\mathbf{X},\mathbf{Y}\in \mathbb{R}^{N\times d}$ where $\mathbf{X}\sim \mu$ and $\mathbf{Y}\sim \nu$, \cite[Lemma 5]{JMLR:v13:gretton12a} characterizes the empirical MMD as
\begin{equation}\label{eq:MMD}
   \text{MMD}_{\sigma}^2(\mathbf{X},\mathbf{Y}):= \frac{1}{N(N-1)}\sum_{n=1}^N \sum_{m\neq n}^N\Big( \gamma_{\sigma}(\mathbf{X}_n,\mathbf{X}_m) + \gamma_{\sigma}(\mathbf{Y}_n,\mathbf{Y}_m)\Big) - \frac{2}{N^2}\sum_{n=1}^N \sum_{m=1}^N \gamma_{\sigma}(\mathbf{X}_n,\mathbf{Y}_m),
\end{equation}
where we denote by $\mathbf{X}_n$ the $n$-th row (or sample) of $\mathbf{X}$ and $\gamma_{\sigma}$ is the Gaussian kernel with bandwidth $\sigma$. An analagous formulation also holds when the data matrices $\mathbf{X}$ and $\mathbf{Y}$ differ in size. Given that the complexity of \eqref{eq:MMD} is  $\mathcal{O}(N^2)$, we instead use the so-called linear-MMD approximation (see \cite[Lemma 14]{JMLR:v13:gretton12a}) when $N$ is large (e.g., $N = 10^4$ in Sections~\ref{subsec:lowdim}~and~\ref{subsec:highdim}). Moreover, due to the sensitivity of the MMD metric on the choice of kernel bandwidth $\sigma$, we follow~\cite{NIPS2009_685ac8ca} by computing the so-called Generalized MMD. This involves evaluating \eqref{eq:MMD} for a finite collection of bandwidths $\mathcal{I} \subseteq \mathbb{R}$ and reporting the maximum value. Throughout our tests, we choose the elements of $\mathcal{I}$ logarithmically spaced between $10^{-2}$ and $10^2$ with $|\mathcal{I}| = 15$. In situations when we compare the accuracy of the generative models at only a single snapshot (see \Cref{subsec:embryo}), we report the Generalized MMD, which we denote by $\text{GMMD}_{\mathcal{I}}^2$. Moreover, when we compare the accuracy of the generative models at several snapshots (see Sections \ref{subsec:lowdim}, \ref{subsec:highdim}, and \ref{subsec:fish}), we report the average of the Generalized MMD across all such snapshots, which we denote by $\smash[b]{\widetilde{\text{GMMD}_{\mathcal{I}}^2}}$. 

\subsection{Low-Dimensional Comparison}\label{subsec:lowdim}

We begin by comparing the speed and accuracy of D-PPMM with both the standard RQ-NSF and Reduced RQ-NSF architectures conditioned on time. Specifically, we will consider inference sample path data $\smash[b]{\{\mathbf{X}^{(t_j)}\}_{j=1}^M}$ arising from the Van der Pol oscillator ($d = 2$), an OU-process ($d = 3$), and the Lorenz-96 system $(d = 4);$ see \Cref{subsec:dataprep}. For these experiments, the terminal time of the Lorenz-96 system is taken as $T = 5$. Given that the problem of interpolation is highly application-dependent, we will first be concerned with quantifying the speed and accuracy of these frameworks as generative models exactly at the measurement times $\smash[b]{\{t_j\}_{j =1}^M}$. 
\begin{figure}[h!]
    \centering
    \includegraphics[width = .9\textwidth]{NewFigures/Figure2.png}
    \caption{\textbf{Low-dimensional qualitative comparison.} We show samples from the D-PPMM and time-conditioned Reduced RQ-NSF generative models in which both models are permitted the same amount of wall-clock training time ($< $ 10 seconds). The coloring shows time-evolution, with blue indicating $t = 0$ and red indicating $t = T.$}
    \label{fig:qual}
\end{figure}

Figure \ref{fig:qual} shows an initial qualitative comparison between D-PPMM and a time-conditioned Reduced RQ-NSF, in which both models were permitted to train for exactly the same amount of wall-clock time. This was achieved by choosing a convergence tolerance $\alpha > 0$, recording the wall-clock time $t_{\alpha}$ after which D-PPMM finishes training, and then training a time-conditioned RQ-NSF for exactly the same amount of time $t_{\alpha}.$ Throughout our comparisons, we do not track the time during which the RQ-NSF framework partitions the training set into mini-batches. In Figure \ref{fig:qual}, we see for each test that D-PPMM produces new samples which appear more representative of the original dynamics and visually have smaller variance than the samples of the Reduced RQ-NSF. For the Van der Pol oscillator and Lorenz-96 system, D-PPMM uses $B =500$ and $L = 0.1$. For the OU-process, D-PPMM uses $B = 2000$ and $L = 0.1$. The RQ-NSFs use a mini-batch size of 100, with the remaining hyperparameter choices detailed in \Cref{subsec:DNSF}.

\begin{figure}[h!]
    \centering
    \includegraphics[width = \textwidth]{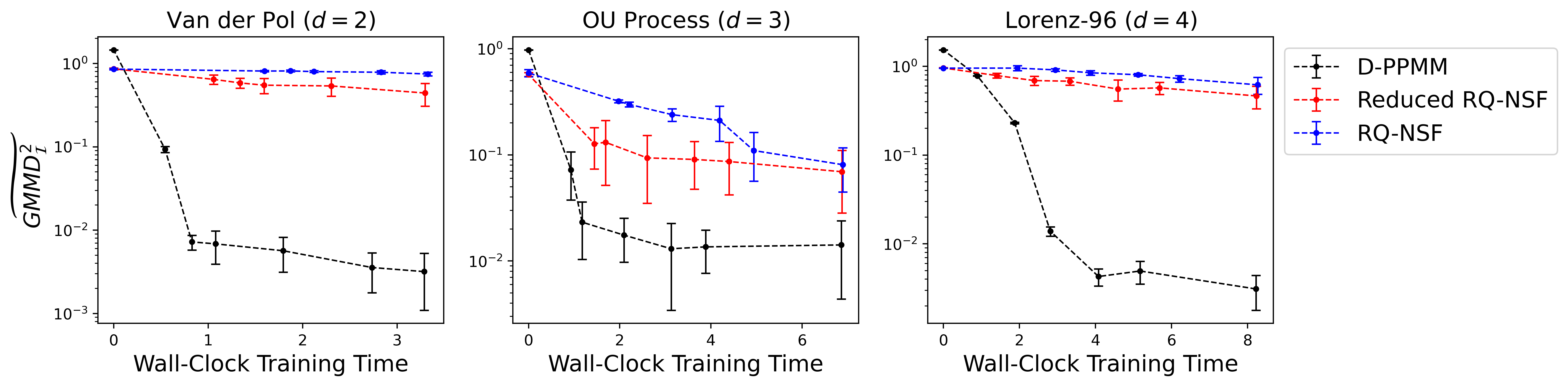}
    \caption{\textbf{Low-dimensional convergence comparison.} The convergence behavior of D-PPMM and the time-conditioned RQ-NSFs as generative models at the observed measurement times for low-dimensional dynamical systems.}
    \label{fig:quant1}
\end{figure}

In Figure \ref{fig:quant1}, we quantitatively compare the efficiency and accuracy of D-PPMM, the time-conditioned RQ-NSF, and time-conditioned Reduced RQ-NSF frameworks. 
Since the time to compute the MMD is quadratic in the number of samples $N=10^4$, the error we use in Figure \ref{fig:quant1} is based on the linear-MMD approximation to the true MMD. In Figure~\ref{fig:quant1}, we find that D-PPMM consistently converges faster and with smaller error than both time-conditioned RQ-NSF frameworks. The hyperparameters used to train the generative models are the same as in Figure \ref{fig:qual}. To account for the randomness during the training, testing, and error quantification processes, we train each model $10$ times with different random seeds and visualize both the mean testing error and the standard deviation over each set of trials for a fixed training time.

\subsection{High-Dimensional Comparison}\label{subsec:highdim}
\begin{figure}[h!]
    \centering
    \includegraphics[width = \textwidth]{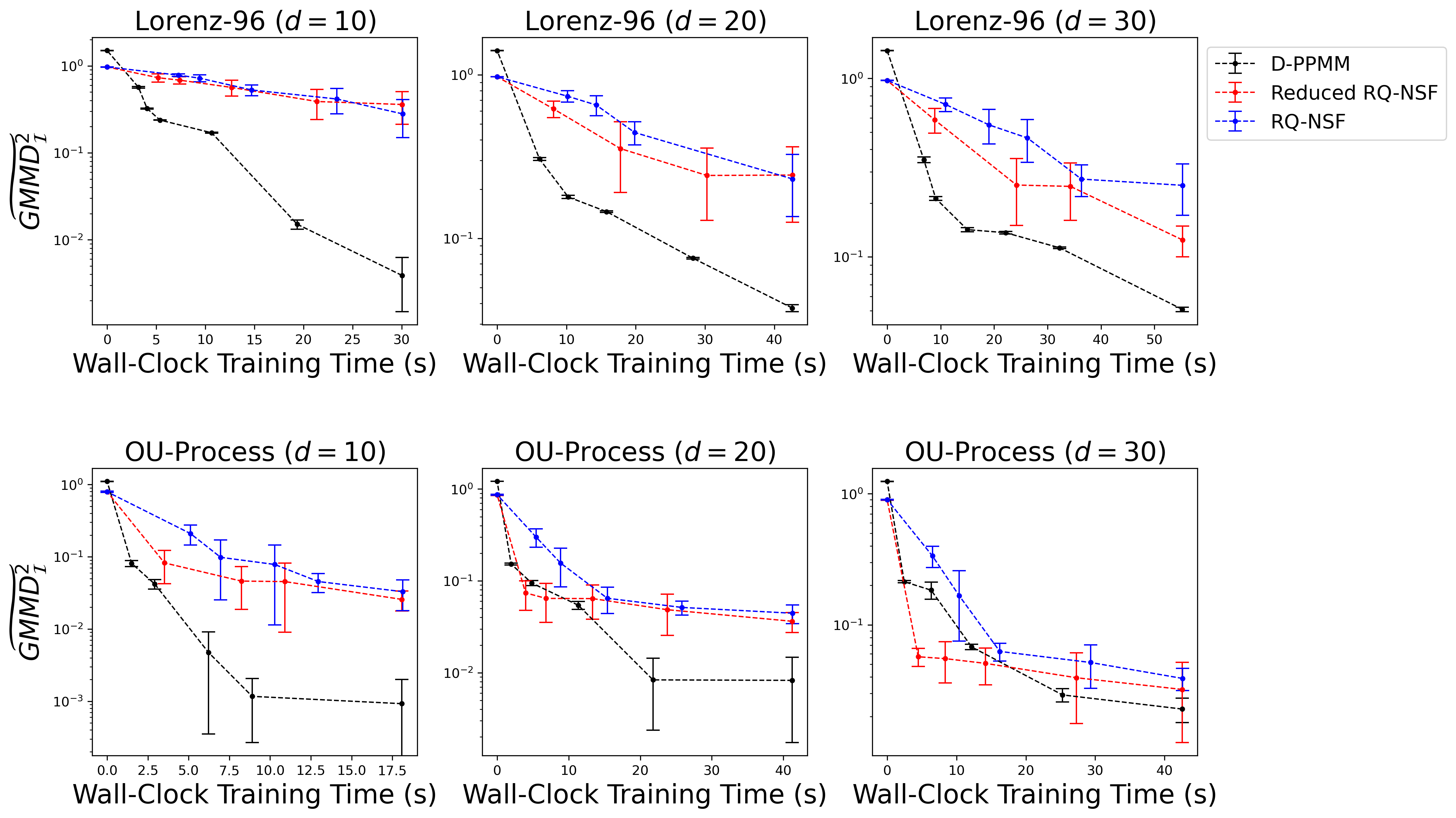}
    \caption{\textbf{High dimensional convergence comparison.} We show the convergence behavior of D-PPMM and the time-conditioned RQ-NSFs for the Lorenz-96 system and OU-process with $d \in \{10,20,30\}$ with $N = 10^4.$ Each model is trained $10$ times with different random seeds for each training time shown. The error is computed according to the linear-MMD approximation.}
    \label{fig:quant2}
\end{figure}

We next turn to the high-dimensional setting and consider both the OU-process and Lorenz-96 system with $d=10$, $20$ and $30$. For these experiments, we set the final time $T = 3.5$ for the Lorenz-96 system. Conducting a similar convergence analysis as in Figure~\ref{fig:quant1}, we find in Figure~\ref{fig:quant2} that D-PPMM remains competitive with the time-conditioned RQ-NSF in the high-dimensional setting. We observe that in these examples, the performances of the time-conditioned RQ-NSFs start to catch up with those from D-PPMM  as the dimension increases.

We remark that the convergence of the D-PPMM algorithm will remain independent of the dimension $d$ only if the underlying density changes along a fixed number of directions as $d$ increases. Due to the $d$-dimensional Brownian noise used in the experiments in Figure~\ref{fig:quant2}, as well as the fact that the first optimal map we learn is from a $d$-dimensional Gaussian, we do not expect the convergence of D-PPMM to be dimension independent. In situations when each component of the state vector of the underlying ODE/SDE evolves independently, such as the OU process, it is possible to decompose the pushforward map along $d$-directions, corresponding to the standard Euclidean basis vectors, which can be learned in parallel without the use of SAVE. This approach relies on prior knowledge about the functional form of the underlying dynamics, and we do not consider this case here. 

\subsection{Fish Schooling Dataset}\label{subsec:fish}
Having studied D-PPMM for synthetic stochastic processes in \Cref{subsec:lowdim,subsec:highdim}, we now consider its application to real-world data sets originating from biological applications. Moreover, instead of comparing D-PPMM and the conditioned RQ-NSF at the observed measurement times, as in \Cref{subsec:lowdim,{subsec:highdim}}, we will now examine their interpolation capabilities. Recall that D-PPMM can interpolate densities using the transport splines algorithm (see \Cref{subsubsec:ts}), whereas the conditional normalizing flow can simply be evaluated at the desired interpolation time (see \Cref{subsec:DNSF}). We begin by studying a dataset of fish schooling behavior \cite{Katz2021-sr}. It features trajectories of 300 golden shiners in a rectangular water tank and contains snapshot data, which records the position $(x_i,y_i) \in \mathbb{R}^2$ of the shiners, sampled at 30 Hz frequency. We consider the first 2301 snapshots of the dataset in our comparison. We also note that the recorded sample size varies per snapshot based on which fish are detectable by the tracking software used \cite{Katz2021-sr}. Thus, it is particularly advantageous that PPMM can be used to learn OT maps between distributions with different sample sizes (see \Cref{subsec:PPMM}). Given the sparsity of observations in this particular dataset, we also find it useful to manually select a relatively large bandwidth when performing the KDE for the 1D projected samples (see \Cref{subsubsec:reg}). 

\begin{figure}[h!]
    \centering
    \includegraphics[width = \textwidth]{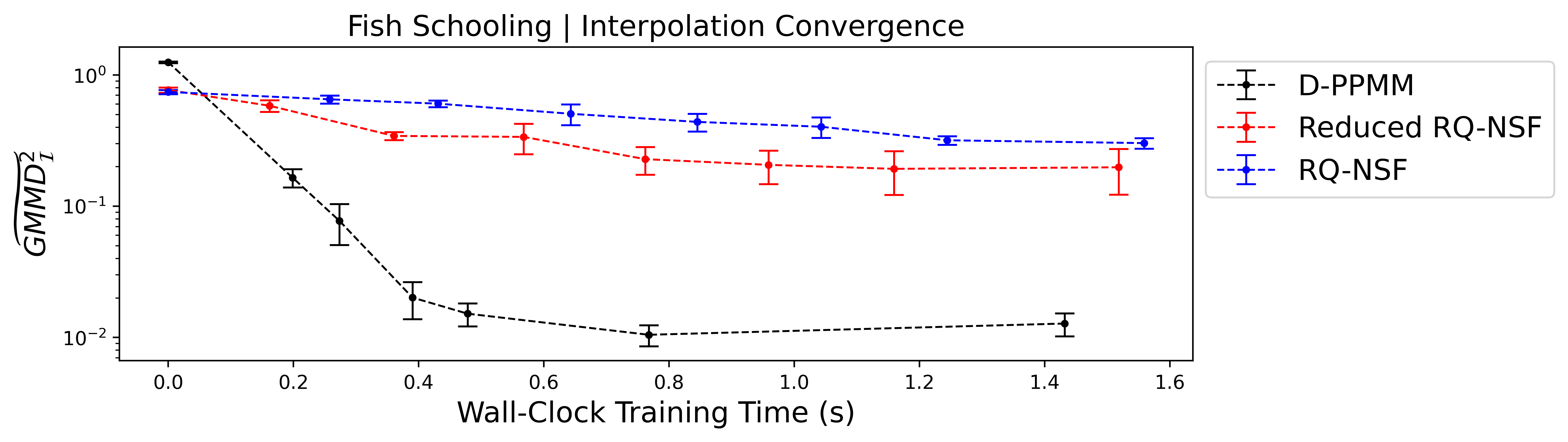}
    \caption{\textbf{Fish schooling convergence comparison.} Error comparison between D-PPMM and the time-conditioned RQ-NSFs at the held-out snapshots $S_2$ as the training time is varied.  }
    \label{fig:fishconv}
\end{figure}

To study the effectiveness of interpolation with D-PPMM and the conditioned RQ-NSF, we construct the generative models only from snapshots with indices in $S_1:=\{100 j : 0\leq j \leq 23\} $ and leave out all other data (note that for this example we index our data starting with 0, rather than 1). This effectively reduces the sampling rate from 30~Hz to  $0.3$~Hz. We then compare the models based on their success at inferring the distribution of golden shiners at the snapshots with indices in $S_2:=\{50,550,1050,1550,2050\}.$ In Figure~\ref{fig:fishconv}, we study the initial convergence behavior of D-PPMM and the conditioned RQ-NSF when inferring the distribution of shiners at the interpolation snapshots $S_2.$ While D-PPMM converges almost instantly, both normalizing flow frameworks make minimal progress during the same training time. These results are consistent with the rapid convergence of D-PPMM displayed in Figure~\ref{fig:quant1}. For the experiment in Figure~\ref{fig:fishconv}, D-PPMM uses a bandwidth of $7.5 \cdot 10^{-2}$, $B = 75$, and $L = 0.25$, and the RQ-NSFs use a mini-batch size of 46. Figure \ref{fig:fishconv} shows the mean and standard deviation of each model's error after training $10$ times with different random seeds. 

\begin{figure}[h!]
    \centering
    \includegraphics[width=\textwidth]{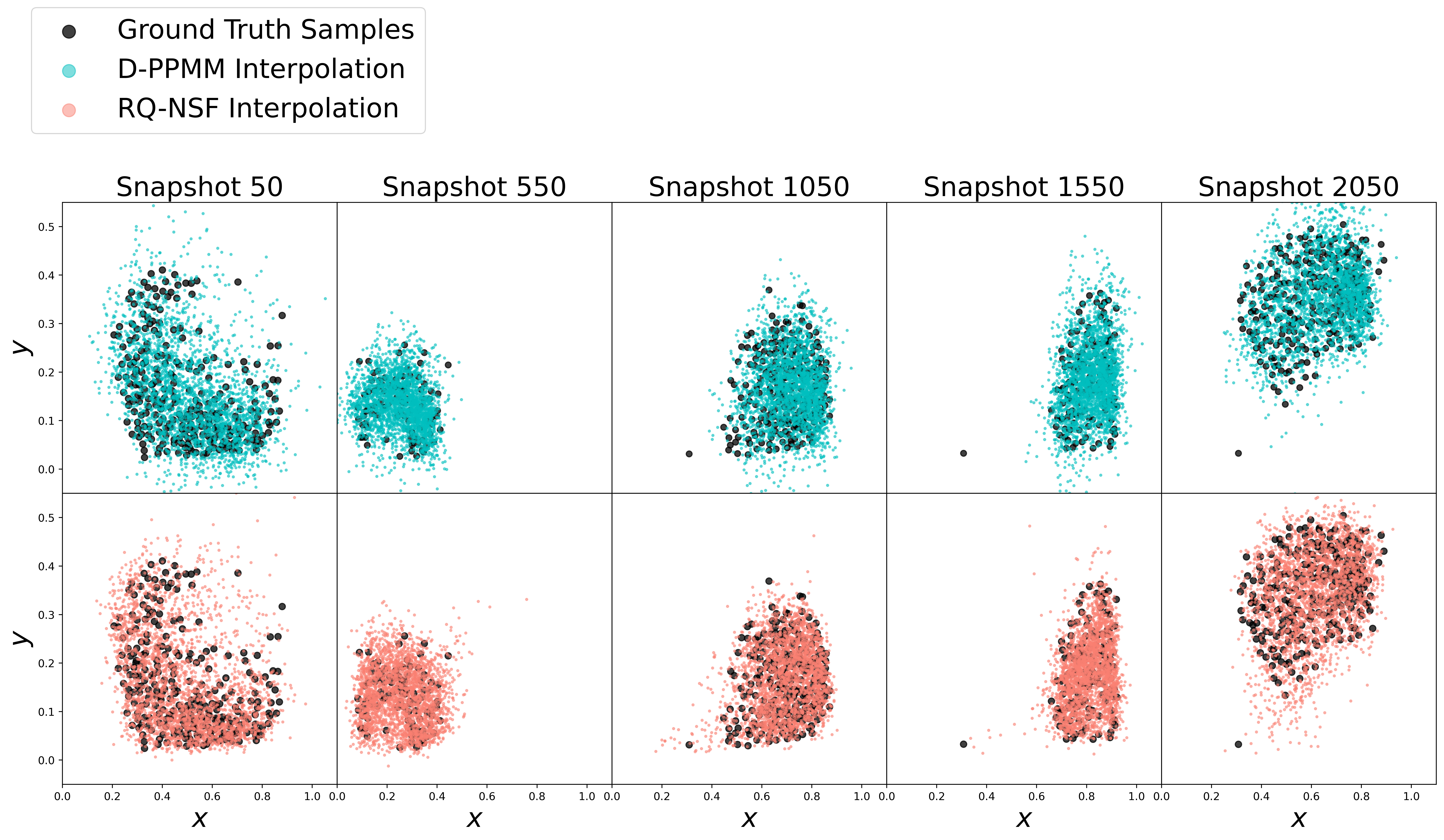}
    \caption{\textbf{Fish schooling sample comparison.} Samples generated by D-PPMM using cubic transport spline interpolation and the conditioned RQ-NSF at the held-out snapshots $S_2$. While D-PPMM trained for less than one second, the RQ-NSF was permitted 1200 seconds of training time.}
    \label{fig:fish_visual}
\end{figure}

While Figure~\ref{fig:fishconv} shows that D-PPMM with transport spline interpolation exhibits rapid initial convergence compared to the conditioned RQ-NSF models, it is also desirable to compare the accuracy of its interpolation with a conditional flow that has been trained for considerably longer than one second. Towards this, in Figure~\ref{fig:fish_visual}, we compare the output of the D-PPMM interpolation with a conditioned RQ-NSF which was trained for 1200 seconds wall-clock time. Note that in Figure \ref{fig:fish_visual} we use the generative models to produce approximately 10$\times$ as many samples as were available during training. For this comparison, the hyperparameters of the models are the same as in Figure \ref{fig:fishconv}. Based on the results in Table~\ref{tab:2}, the use of an RQ-NSF results in more accurate interpolation for this example, but also requires significantly more training time. On the other hand, D-PPMM can still produce reasonable interpolants with several orders of magnitude less training time (about $1/1500$ of what the RQ-NSF required).

\begin{table}[h!]
    \centering
      \begin{tabular}{|c|c|c|c|c|c|c|c|c|c|c|}
    \hline
      Method & Training Time (s) & $\widetilde{\text{GMMD}^2_{\mathcal{I}}}$ $(\times 10^{-3}$)    \\
      \hline
     D-PPMM & \textbf{0.81} & $13.51\pm 4.35$ \\ 
     Reduced RQ-NSF & 0.81 & $319.25 \pm 5.49$ \\
     RQ-NSF & 1200 & $\mathbf{8.76 \pm 0.85}$ \\     \hline
    \end{tabular}
    \vspace{.2cm}
    \caption{Quantitative comparison of D-PPMM and the conditioned RQ-NSF at the held-out snapshots $S_2$. The average and standard deviation of the errors are computed over ten different simulations, while holding the same training seed fixed.}
    \label{tab:2}
\end{table}
\subsection{Embryoid Body Dataset}\label{subsec:embryo}
\begin{figure}
  \centering
 \subfloat[Inference data; the third snapshot is left out of the training set for testing.]{ \includegraphics[height=3.5cm]{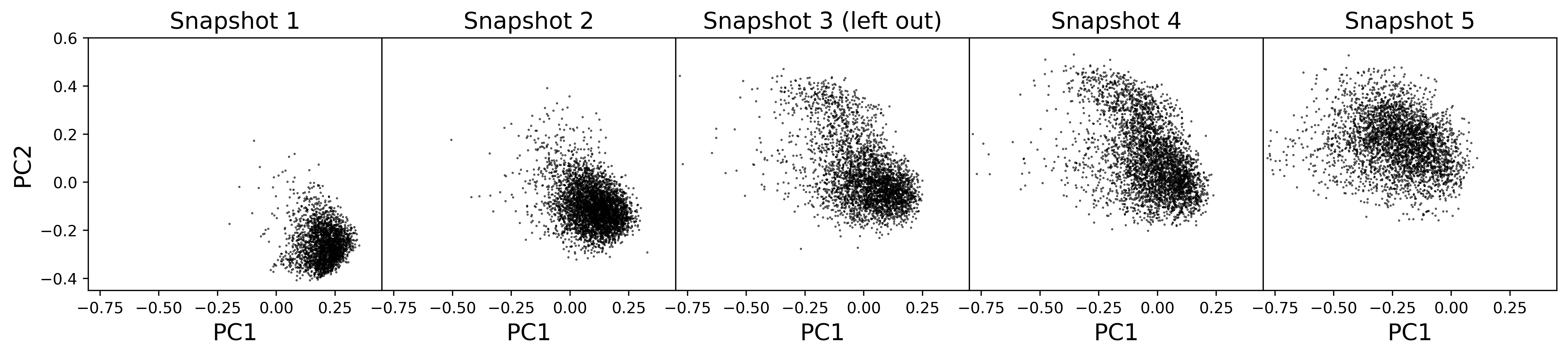}\label{fig:y equals x}}\\
 \subfloat[D-PPMM transport splines]{ \includegraphics[height=3.5cm]{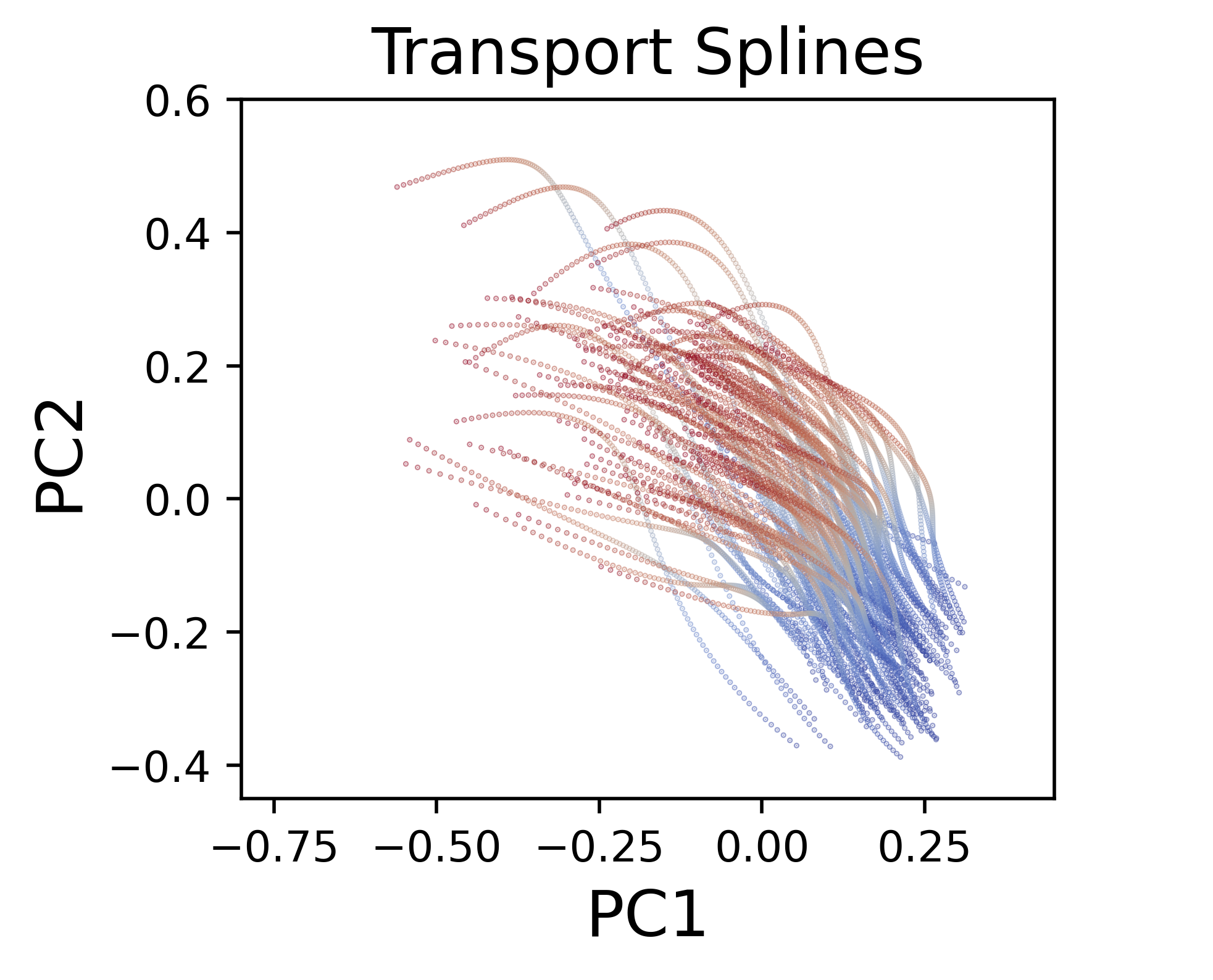}}
 \hspace{.75cm}
  \subfloat[Interpolation results for D-PPMM and RQ-NSFs with various training times]{ \includegraphics[height=3.5cm]{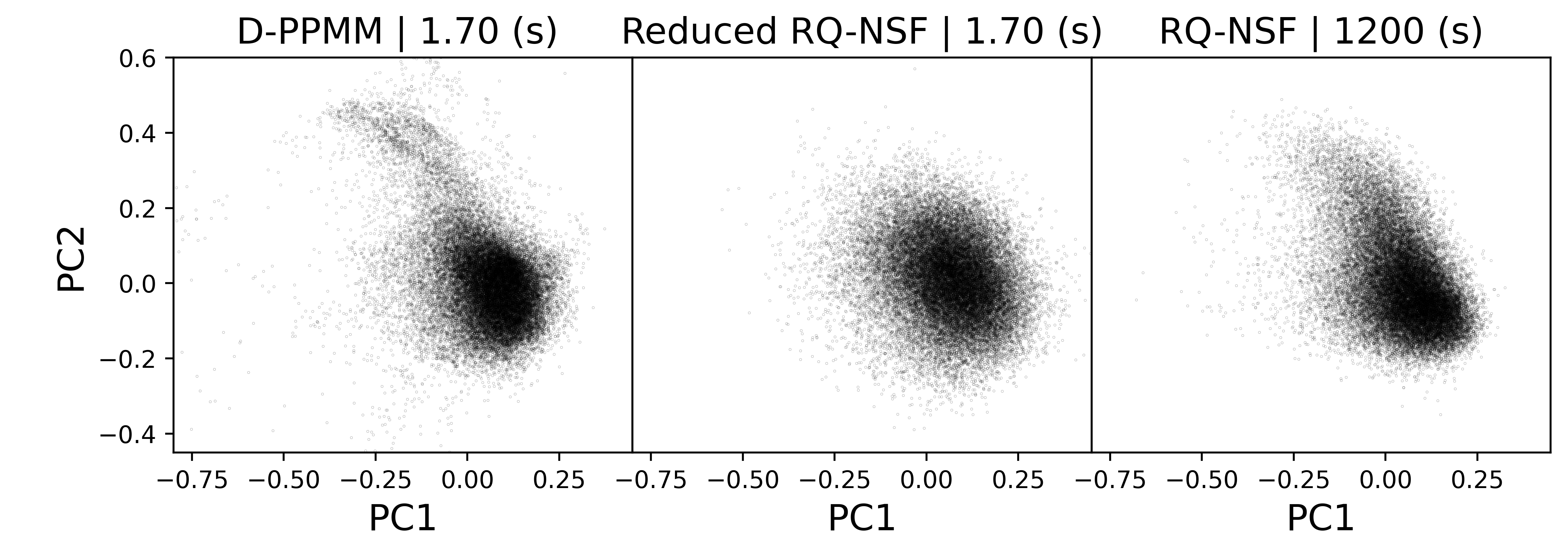}    \label{fig:three sin x}}
\caption{\textbf{Embryoid body dataset interpolation comparison.} Top row: embryoid body training and testing datasets. Bottom left: Transport splines that D-PPMM uses to interpolate the left-out snapshot. Bottom right: interpolation comparison of the third snapshot using D-PPMM with cubic transport spline interpolation and a time-conditioned RQ-NSF. We show results for 1.70 and 1200 seconds of wall-clock training time.}
\label{fig:emb}
\end{figure}

Next, we study cellular trajectories from an embryoid body dataset~\cite{Moon2018-ld}. We remark that this embryoid dataset is used as an example in~\cite[Figs.~7-8]{https://doi.org/10.48550/arxiv.2002.04461}. The dataset features $5$ evenly-spaced snapshots of a single cell RNA sequencing over a $27$-day period with approximately $3\cdot 10^4$ cells in total. We access a pre-processed version of the dataset from~\cite{https://doi.org/10.48550/arxiv.2002.04461}, which filters out rare genes and dead cells. This dataset contains approximately $1.6\cdot 10^4$ cells and is projected onto its first $5$ principal components. The time evolution of these samples is shown in the top row of Figure~\ref{fig:emb}. Similar to \cite{{https://doi.org/10.48550/arxiv.2002.04461}}, we leave the third snapshot out from training and attempt to infer it from the remaining $4$ snapshots. 

\begin{table}[h!]
    \centering
    \begin{tabular}{|c|c|c|c|c|c|c|c|c|c|c|}
    \hline
      Method & Training Time (s) & $\text{GMMD}^2_{\mathcal{I}}$ $(\times 10^{-3}$)    \\
      \hline
     D-PPMM & $\mathbf{1.70}$ &$24.97 \pm 1.26$ \\ 
     Reduced RQ-NSF & 1.70 & $54.99 \pm 2.28$ \\
     RQ-NSF & 1200 & $\mathbf{24.14 \pm 1.04}$ \\     \hline
    \end{tabular}
    \vspace{.2cm}
    \caption{Quantitative comparison of D-PPMM and the conditioned RQ-NSF at the held-out snapshot 3. The average and standard deviation of the error are computed over 10 generations with the same training seed.}
    \label{tab:3}
\end{table}

The samples generated by D-PPMM and the conditioned RQ-NSF are shown in the bottom row of Figure~\ref{fig:emb}. Note that for this example, D-PPMM finished training with a convergence tolerance $\alpha = 10^{-5}$, $B = 500$, and $L = 0.1$ after 1.70 seconds. For comparison, we show the progress that the Reduced RQ-NSF makes after the same amount of time, as well as the samples of an RQ-NSF which has been allotted 1200 seconds of training time. For this experiment, the RQ-NSFs use a mini-batch size of 92 samples. We note that in this case, the time-conditioned RQ-NSF converged with comparable error with D-PPMM with the transport spline interpolation, but that D-PPMM required several orders of magnitude less training time (about $1/700$ of what the RQ-NSF required); see Table \ref{tab:3}.

\section{Conclusions}\label{sec:conclusions}
We have introduced D-PPMM, which constructs a generative model of the sample paths of an evolving density by coupling observed snapshots via OTMs and subsequently applying transport spline interpolation. Though the concept of joining samples from successive snapshots via OTMs is not new, we have shown that the proposed approach is a computationally practical option that can circumvent the large neural network architectures associated with many popular generative models of dynamical systems. Specifically, we have conducted a thorough comparison of the convergence behavior of D-PPMM with state-of-the-art RQ-NSFs conditioned on time.
In many of the examples we studied, D-PPMM was more efficient and accurate than the time-conditioned RQ-NSFs. As the dimension $d$ was increased, the difference in performance between the generative models became less noticeable, but D-PPMM still remained competitive. We remark that in our computations in \Cref{sec:experiments}, we did not parallelize the training procedure for D-PPMM, but that doing so may lead to even faster performance. More specifically, the computation of the optimal map between two successive snapshots does not affect the computation of the optimal map between two different successive snapshots. Thus, the training for-loop in Algorithm \ref{alg:D-PPMM} naturally lends itself to parallelization.

While we have shown that the D-PPMM method is competitive with the conditional normalizing flow in several settings, there are also situations in which the use of the normalizing flow may be preferable. In particular, D-PPMM can suffer from error accumulation when a large number of snapshots are considered, whereas the normalizing flow does not encounter this problem; see Figure \ref{fig:F1}. Furthermore, the normalizing flow can be trained on irregularly sampled data, while D-PPMM is expected to struggle when there are relatively few samples at a given observation time. Lastly, the normalizing flow offers additional flexibility for modeling conditional probabilities, as one can use the statistics of previous timesteps as conditioning factors and the sampling time itself to model the density at future timesteps. This feature is expected to be helpful when one wishes to predict the dynamics of a stochastic process rather than interpolate between known measurement times, as in \cite{rasul2021multivariate}. Future work may focus on extending D-PPMM to remain competitive with the normalizing flow in these settings.

\section*{Author Declarations}
\subsection*{Conflict of Interest}
 The authors have no conflicts to disclose.
\subsection*{Author Contributions}
\textbf{Jonah Botvinick-Greenhouse}: Formal analysis (lead), methodology (equal), writing -- original draft preparation (lead), writing -- review and editing (equal). \textbf{Yunan Yang}: Methodology (equal), writing -- original draft preparation (supporting), writing -- review and editing (equal). \textbf{Romit Maulik}: Conceptualization (lead), methodology (equal), writing -- original draft preparation (supporting), writing -- review and editing (equal).
\subsection*{Data Availability}

The data used in Sections \ref{subsec:lowdim} and \ref{subsec:highdim} is available from the authors upon reasonable request. The data used in \Cref{subsec:fish} is openly available through \cite{Katz2021-sr}, and the data used in \Cref{subsec:embryo} is openly available through \cite{Moon2018-ld} and \cite{https://doi.org/10.48550/arxiv.2002.04461}.
\section*{Acknowledgements}
 This work was supported by the U.S. Department of Energy (DOE), Office of Science, Office of Advanced Scientific Computing Research (ASCR), under Contract No. DEAC02–06CH11357, at Argonne National Laboratory. We also acknowledge funding support from ASCR for DOE-FOA-2493, ``Data-intensive scientific machine learning''. This research was supported in part by an appointment with the National Science Foundation (NSF) Mathematical Sciences
Graduate Internship (MSGI) Program sponsored by the NSF Division of Mathematical Sciences. This program is administered by
the Oak Ridge Institute for Science and Education (ORISE) through an interagency agreement between the U.S. Department of
Energy (DOE) and NSF. ORISE is managed for DOE by ORAU. All opinions expressed in this paper are the author's and do not
necessarily reflect the policies and views of NSF, ORAU/ORISE, or DOE. This paper was supported in part by a fellowship award under contract FA9550-21-F-0003 through the National Defense Science and Engineering Graduate (NDSEG) Fellowship Program, sponsored by the Air Force Research Laboratory (AFRL), the Office of Naval Research (ONR) and the Army Research Office (ARO). Y.~Yang acknowledges support from Dr.~Max R\"ossler, the Walter Haefner Foundation, and the ETH Z\"urich Foundation.

\bibliographystyle{plain}  
\bibliography{references.bib}

\begin{thebibliography}{10}

\bibitem{arjovsky2017wasserstein}
Martin Arjovsky, Soumith Chintala, and L{\'e}on Bottou.
\newblock Wasserstein generative adversarial networks.
\newblock In {\em International conference on machine learning}, pages
  214--223. PMLR, 2017.

\bibitem{benamou2019second}
Jean-David Benamou, Thomas~O Gallou{\"e}t, and Fran{\c{c}}ois-Xavier Vialard.
\newblock Second-order models for optimal transport and cubic splines on the
  wasserstein space.
\newblock {\em Foundations of Computational Mathematics}, 19:1113--1143, 2019.

\bibitem{10.1214/10-AOS799}
Z.~I. Botev, J.~F. Grotowski, and D.~P. Kroese.
\newblock {Kernel density estimation via diffusion}.
\newblock {\em The Annals of Statistics}, 38(5):2916 -- 2957, 2010.

\bibitem{both2019temporal}
Gert-Jan Both and Remy Kusters.
\newblock Temporal normalizing flows.
\newblock {\em arXiv preprint arXiv:1912.09092}, 2019.

\bibitem{NEURIPS2018_69386f6b}
Ricky T.~Q. Chen, Yulia Rubanova, Jesse Bettencourt, and David~K Duvenaud.
\newblock Neural ordinary differential equations.
\newblock In S.~Bengio, H.~Wallach, H.~Larochelle, K.~Grauman, N.~Cesa-Bianchi,
  and R.~Garnett, editors, {\em Advances in Neural Information Processing
  Systems}, volume~31. Curran Associates, Inc., 2018.

\bibitem{https://doi.org/10.48550/arxiv.2010.12101}
Sinho Chewi, Julien Clancy, Thibaut~Le Gouic, Philippe Rigollet, George
  Stepaniants, and Austin~J. Stromme.
\newblock {Fast and smooth interpolation on Wasserstein space}, 2020.

\bibitem{deng2021continuous}
Ruizhi Deng, Marcus~A Brubaker, Greg Mori, and Andreas Lehrmann.
\newblock Continuous latent process flows.
\newblock In A.~Beygelzimer, Y.~Dauphin, P.~Liang, and J.~Wortman Vaughan,
  editors, {\em Advances in Neural Information Processing Systems}, 2021.

\bibitem{doi:10.1080/03610920008832598}
R.~Dennis~Cook.
\newblock {SAVE: a method for dimension reduction and graphics in regression}.
\newblock {\em Communications in statistics-Theory and methods},
  29(9-10):2109--2121, 2000.

\bibitem{https://doi.org/10.48550/arxiv.1906.04032}
Conor Durkan, Artur Bekasov, Iain Murray, and George Papamakarios.
\newblock Neural spline flows.
\newblock {\em Advances in neural information processing systems}, 32, 2019.

\bibitem{nflows}
Conor Durkan, Artur Bekasov, Iain Murray, and George Papamakarios.
\newblock {nflows}: normalizing flows in {PyTorch}, November 2020.

\bibitem{feng2021solving}
Xiaodong Feng, Li~Zeng, and Tao Zhou.
\newblock Solving time dependent {Fokker-Planck} equations via temporal
  normalizing flow.
\newblock {\em arXiv preprint arXiv:2112.14012}, 2021.

\bibitem{friedman1981projection}
Jerome~H Friedman and Werner Stuetzle.
\newblock Projection pursuit regression.
\newblock {\em Journal of the American statistical Association},
  76(376):817--823, 1981.

\bibitem{NIPS2009_685ac8ca}
Kenji Fukumizu, Arthur Gretton, Gert Lanckriet, Bernhard Sch\"{o}lkopf, and
  Bharath~K. Sriperumbudur.
\newblock Kernel choice and classifiability for rkhs embeddings of probability
  distributions.
\newblock In Y.~Bengio, D.~Schuurmans, J.~Lafferty, C.~Williams, and
  A.~Culotta, editors, {\em Advances in Neural Information Processing Systems},
  volume~22. Curran Associates, Inc., 2009.

\bibitem{golightly2010markov}
Andrew Golightly and Darren~J Wilkinson.
\newblock {Markov chain Monte Carlo algorithms for SDE parameter estimation}.
\newblock {\em Learning and Inference for Computational Systems Biology}, pages
  253--276, 2010.

\bibitem{goodfellow2020generative}
Ian Goodfellow, Jean Pouget-Abadie, Mehdi Mirza, Bing Xu, David Warde-Farley,
  Sherjil Ozair, Aaron Courville, and Yoshua Bengio.
\newblock Generative adversarial networks.
\newblock {\em Communications of the ACM}, 63(11):139--144, 2020.

\bibitem{JMLR:v13:gretton12a}
Arthur Gretton, Karsten~M. Borgwardt, Malte~J. Rasch, Bernhard Sch{{\"o}}lkopf,
  and Alexander Smola.
\newblock A kernel two-sample test.
\newblock {\em Journal of Machine Learning Research}, 13(25):723--773, 2012.

\bibitem{jia2019neural}
Junteng Jia and Austin~R Benson.
\newblock Neural jump stochastic differential equations.
\newblock {\em Advances in Neural Information Processing Systems}, 32, 2019.

\bibitem{Katz2021-sr}
Y~Katz, K~Tunstr{\o}m, C~Ioannou, C~Huepe, and I~Couzin.
\newblock {\em Fish Schooling Data Subset}.
\newblock Oregon State University, 2021.

\bibitem{kidger2021neural}
Patrick Kidger, James Foster, Xuechen Li, and Terry~J Lyons.
\newblock {Neural SDEs as infinite-dimensional GANS}.
\newblock In {\em International Conference on Machine Learning}, pages
  5453--5463. PMLR, 2021.

\bibitem{Kobyzev_2021}
Ivan Kobyzev, Simon~JD Prince, and Marcus~A Brubaker.
\newblock Normalizing flows: An introduction and review of current methods.
\newblock {\em IEEE transactions on pattern analysis and machine intelligence},
  43(11):3964--3979, 2020.

\bibitem{liu2019neural}
Xuanqing Liu, Tesi Xiao, Si~Si, Qin Cao, Sanjiv Kumar, and Cho-Jui Hsieh.
\newblock {Neural SDE: Stabilizing neural ODE networks with stochastic noise}.
\newblock {\em arXiv preprint arXiv:1906.02355}, 2019.

\bibitem{Lu_2022}
Yubin Lu, Romit Maulik, Ting Gao, Felix Dietrich, Ioannis~G. Kevrekidis, and
  Jinqiao Duan.
\newblock Learning the temporal evolution of multivariate densities via
  normalizing flows.
\newblock {\em Chaos: An Interdisciplinary Journal of Nonlinear Science},
  32(3):033121, mar 2022.

\bibitem{mbalawata2013parameter}
Isambi~S Mbalawata, Simo S{\"a}rkk{\"a}, and Heikki Haario.
\newblock {Parameter estimation in stochastic differential equations with
  Markov chain Monte Carlo and non-linear Kalman filtering}.
\newblock {\em Computational Statistics}, 28(3):1195--1223, 2013.

\bibitem{https://doi.org/10.48550/arxiv.2106.05838}
Cheng Meng, Yuan Ke, Jingyi Zhang, Mengrui Zhang, Wenxuan Zhong, and Ping Ma.
\newblock Large-scale optimal transport map estimation using projection
  pursuit.
\newblock {\em Advances in Neural Information Processing Systems}, 32, 2019.

\bibitem{Moon2018-ld}
Kevin Moon.
\newblock Embryoid body data for {PHATE}, 2018.

\bibitem{NIELSEN200083}
Jan~Nygaard Nielsen, Henrik Madsen, and Peter~C. Young.
\newblock {Parameter estimation in stochastic differential equations: An
  overview}.
\newblock {\em Annual Reviews in Control}, 24:83--94, 2000.

\bibitem{tommy_odland_2018_2392268}
Tommy Odland.
\newblock tommyod/kdepy: Kernel density estimation in { Python}, December 2018.

\bibitem{SP}
Grigorios~A. Pavliotis.
\newblock {\em Stochastic Processes and Applications}.
\newblock Springer New York, NY, 2014.

\bibitem{peyre2019computational}
Gabriel Peyr{\'e}, Marco Cuturi, et~al.
\newblock Computational optimal transport: With applications to data science.
\newblock {\em Foundations and Trends{\textregistered} in Machine Learning},
  11(5-6):355--607, 2019.

\bibitem{rasul2021multivariate}
Kashif Rasul, Abdul-Saboor Sheikh, Ingmar Schuster, Urs~M Bergmann, and Roland
  Vollgraf.
\newblock Multivariate probabilistic time series forecasting via conditioned
  normalizing flows.
\newblock In {\em International Conference on Learning Representations}, 2021.

\bibitem{rout2021generative}
Litu Rout, Alexander Korotin, and Evgeny Burnaev.
\newblock Generative modeling with optimal transport maps.
\newblock {\em arXiv preprint arXiv:2110.02999}, 2021.

\bibitem{salimans2015markov}
Tim Salimans, Diederik Kingma, and Max Welling.
\newblock {Markov chain Monte-Carlo and variational inference: Bridging the
  gap}.
\newblock In {\em International conference on machine learning}, pages
  1218--1226. PMLR, 2015.

\bibitem{scholler2021flomo}
Christoph Sch{\"o}ller and Alois Knoll.
\newblock Flomo: Tractable motion prediction with normalizing flows.
\newblock In {\em 2021 IEEE/RSJ International Conference on Intelligent Robots
  and Systems (IROS)}, pages 7977--7984. IEEE, 2021.

\bibitem{scott2012multivariate}
David~W Scott.
\newblock Multivariate density estimation and visualization.
\newblock {\em Handbook of computational statistics: Concepts and methods},
  pages 549--569, 2012.

\bibitem{song2020score}
Yang Song, Jascha Sohl-Dickstein, Diederik~P Kingma, Abhishek Kumar, Stefano
  Ermon, and Ben Poole.
\newblock Score-based generative modeling through stochastic differential
  equations.
\newblock {\em arXiv preprint arXiv:2011.13456}, 2020.

\bibitem{https://doi.org/10.48550/arxiv.2002.04461}
Alexander Tong, Jessie Huang, Guy Wolf, David van Dijk, and Smita Krishnaswamy.
\newblock Trajectorynet: A dynamic optimal transport network for modeling
  cellular dynamics, 2020.

\bibitem{villani2009optimal}
C{\'e}dric Villani.
\newblock {\em Optimal transport: old and new}, volume 338.
\newblock Springer, 2009.

\bibitem{wiese2020quant}
Magnus Wiese, Robert Knobloch, Ralf Korn, and Peter Kretschmer.
\newblock Quant gans: deep generation of financial time series.
\newblock {\em Quantitative Finance}, 20(9):1419--1440, 2020.

\bibitem{winkler}
Christina Winkler, Daniel Worrall, Emiel Hoogeboom, and Max Welling.
\newblock Learning likelihoods with conditional normalizing flows.
\newblock {\em arXiv preprint arXiv:1912.00042}, 2019.

\bibitem{yang2020physics}
Liu Yang, Dongkun Zhang, and George~Em Karniadakis.
\newblock Physics-informed generative adversarial networks for stochastic
  differential equations.
\newblock {\em SIAM Journal on Scientific Computing}, 42(1):A292--A317, 2020.

\bibitem{YANG2020110026}
Xiangfeng Yang, Yuhan Liu, and Gyei-Kark Park.
\newblock Parameter estimation of uncertain differential equation with
  application to financial market.
\newblock {\em Chaos, Solitons \& Fractals}, 139:110026, 2020.

\end{thebibliography}


\end{document}